%% file: hope_prepare.tex
\documentclass[letterpaper, 10pt, conference]{IEEEtran}

\IEEEoverridecommandlockouts
\usepackage{cite}
\usepackage{amsmath,amssymb,amsfonts}
\usepackage{algorithm}
\usepackage[noend]{algpseudocode}
\usepackage{graphicx}
\usepackage{textcomp}
\usepackage{xcolor}
\usepackage[inkscapelatex=false]{svg}
\usepackage{tikz}
\usepackage{balance}
\usetikzlibrary{shapes.geometric,positioning}
\newcommand{\ego}{\text{ego}}
\newcommand{\env}{\text{env}}
\newcommand{\hope}{\text{hope}}
\newcommand{\safe}{\text{safe}}
\newcommand{\seen}{\text{seen}}
\newcommand{\hid}{\text{hid}}
\newcommand{\FoV}{\mathcal{F}o\mathcal{V}}
\newcommand{\Occ}{\mathcal{O}cc}
\newcommand{\Reach}{\mathcal{R}}
\algrenewcommand\algorithmicrequire{\textbf{Input:}}
\algrenewcommand\algorithmicensure{\textbf{Output:}}

\begin{document}
\setlength{\textfloatsep}{1.0em}
\addtolength{\topmargin}{0.2in}

\title{\vspace{0.1in} Hope for the Best, Prepare for the Worst: \\ Occlusion-Aware Contingency Planning for \\ Autonomous Vehicles
\thanks{$^{1}$ KTH Royal Institute of Technology, $^{2}$ Traton AB, 
$^{3}$ Chalmers University of Technology, $^{4}$ Univeristy of Gothenburg.

Emails: \{trulsny, tumova\}@kth.se, anna.gautier@chalmers.se. 
The work was partially supported by the Wallenberg AI, Autonomous Systems and  Software  Program (WASP) funded by the Knut and Alice Wallenberg Foundation.}
}

\author{Truls Nyberg$^{1,2}$, Anna Gautier$^{3,4}$, Jana Tumova$^{1}$}

\maketitle

\begin{abstract}
The deployment of autonomous vehicles in urban environments introduces significant safety challenges, particularly in scenarios with occlusions, where critical traffic participants may be hidden from view. Recent accidents involving driverless vehicles highlight the importance of motion planners that explicitly addresses the risks posed by occlusions.
In this work, we propose a formal, occlusion-aware trajectory planning framework that guarantees collision avoidance even when there are possible hidden traffic participants. Building on our previous methods that apply reachability analysis to sequentially determine the possible states of hidden traffic participants, we integrate a tree-based motion planner capable of reasoning over future observations and the absence thereof. This approach reduces conservativeness while maintaining safety guarantees.
We demonstrate the effectiveness of our framework in a challenging simulated occluded scenario, showing that it pro-actively and efficiently guarantees collision-avoidance. 
\end{abstract}

\section{Introduction}
\begin{figure}
    \centering
    \includegraphics[width=\linewidth]{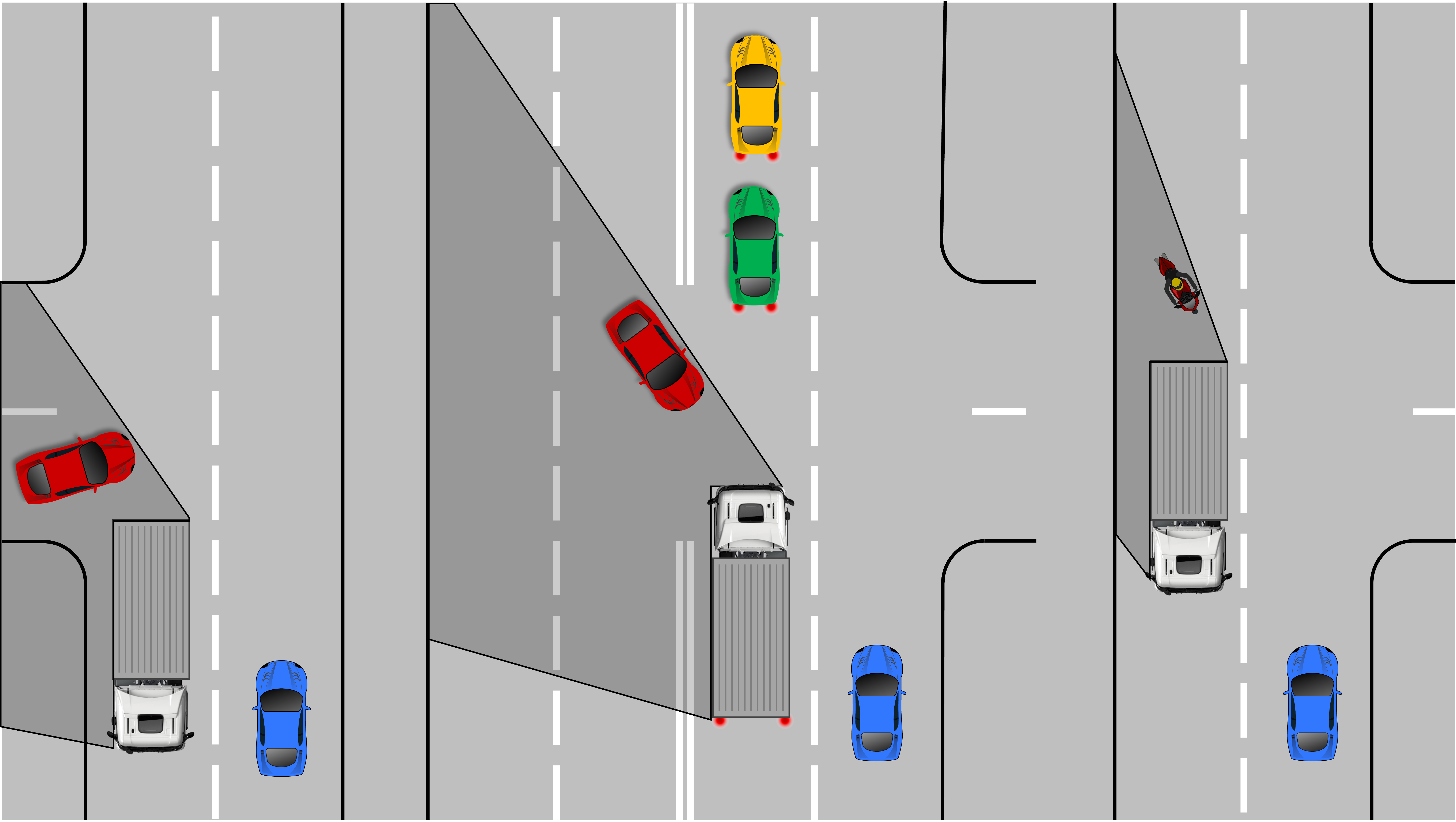}
    \caption{Illustration of three intersection scenarios inspired by recent real-world AV accidents, highlighting critical occlusions caused by large vehicles (e.g., trucks). Regardless of specific right-of-way rules, all traffic participants—including the blue AV—retain a duty of care and must exercise caution.}
    \label{fig:examples}
\end{figure}

We focus on the challenging task of planning efficient, collision-free trajectories for autonomous vehicles (AVs) in scenarios with limited visibility due to occlusions and sensor obstructions. Fig.~\ref{fig:examples} presents three illustrative scenarios inspired by real-world AV accidents where limited visibility led to collisions with suddenly emerging traffic participants. Such scenarios are practically important; for instance, Apple's 2023 disengagement report to the California Department of Motor Vehicles described 11 instances where ``the test vehicle sensor field of view was obstructed, preventing the adequate perception of the surrounding environment and requiring disengagement by the Safety Driver''~\cite{dmv2023}.

Accidents involving driverless vehicles operating under deployment permits in the US further underscore the challenge. In October 2023, Waymo submitted a collision report to the National Highway Traffic Safety Administration (NHTSA) after its vehicle collided with a passenger car occluded by a city bus at an intersection in San Francisco. A similar accident occurred in February 2024, when a Waymo vehicle struck a cyclist occluded by a delivery truck, also at a San Francisco intersection. In April 2024, a third collision was reported involving a Waymo vehicle in Phoenix, where the view was obstructed by an SUV in the adjacent lane stopped before an intersection, preventing detection of a left-turning vehicle crossing the Waymo vehicle's path. Although only minor injuries were reported, the collisions underscore the difficulty of these scenarios. The reports assign no blame and merely state facts, but since neither autonomous nor human-driven vehicles managed to avoid collisions, the situations were evidently challenging for both~\cite{nhtsa2023}.

However, realizing the safety benefits of autonomous vehicles requires formally addressing occlusions. Specifically, planned trajectories must provably avoid collisions with hidden but rule-compliant traffic participants. A key challenge is maintaining safety guarantees without overly conservative driving. Existing approaches~\cite{ftu, ptu, Nager2019}, address this by explicitly tracking possible states of hidden participants, limiting their feasible trajectories and thus the areas the ego vehicle must avoid to guarantee collision-free driving.

Building on this foundation, we now show how the AV's driving performance can be increased even further by incorporating this formal approach into a tree-based occlusion-aware motion planner. This planner not only reasons over future observations but also accounts for the absence of information, enabling more efficient and effective trajectory planning. Additionally, we identify specific challenges that can arise in occlusion-aware planning and provide strategies to address them, while offering an outlook on how learning-based planners can handle rare and safety-critical scenarios such as those involving occlusions.

\section{Related Work}
Safe motion planning for AVs requires verifying that planned trajectories remain collision-free, even when other agents may be partially observed or entirely hidden. This section surveys three key areas of prior work that inform our approach: (i) safety filters for verifying planned trajectories; (ii) occlusion-aware decision-making for handling hidden agents; and (iii) contingency- and policy-based planning under incomplete information. We conclude by positioning our approach as a fusion of these strands, enabling safe and efficient motion planning in the presence of occlusions.

The safety case for an AV typically builds upon several safety arguments, supported by evidence from collected data~\cite{koopman2023}.
To argue that the motion planner yields a feasible and collision free plan, a safety filter can be applied. 
In~\cite{hsu2024}, the authors present a unified view of safety-critical control in autonomous systems. 
A safety filter monitors the motion planner and intervenes with a fallback policy if needed. 
The monitor can check the planned trajectory for collisions under some worst-case assumptions of other road users behaviors, as in~\cite{althoff2014}, or employ simpler safety specifications, such as the safe distances defined by the Responsibility-Sensitive Safety (RSS) model~\cite{RSS}. 
The fallback policy can be an optimized fail-safe trajectory, as proposed in~\cite{Pek2021}, or an action classified as a \emph{proper response} according to the RSS model.

To handle occlusions, the monitor or  worst-case predictior must consider possible hidden traffic participants that are not visible from the AV's sensors~\cite{Orzechowski2018, Nager2019, Koschi2021, Trauth2023}.
It has also been shown that AVs benefit from tracking and bounding possible positions~\cite{ftu, Nager2019} and velocities~\cite{ptu}. Similarly, prior work demonstrates how vehicle-to-everything communication can be utilized if accessible to share information about unseen areas~\cite{stu}. 
The motion planner's efficiency can be improved by reducing the conservative state estimates for the hidden participants. 
However, without a filter-aware plan, the collision avoidance gauntness may still be achieved with an uncomfortable or sub-optimal intervention~\cite{hsu2024}.

In~\cite{Hu2022}, the authors propose formulating a stochastic optimal control problem to achieve a so-called \emph{shielding-aware} robust plan. The approach is extended to also consider the interactive behavior of another agent in~\cite{Hu2024}. 
However, the method only handles seen traffic participants and scales poorly both with the number of agents, and with increasing planning horizons.

The occlusion-aware planning problem can also be formulated as a Partially Observable Markov Decision Process (POMDP), as done in~\cite{Hubmann2019} and \cite{bouton_scalable_2018}. 
However, POMDPs with continuous state and action spaces suffer from the curse of dimensionality. 
Online POMDP solvers  such as~\cite{Ye2017} or~\cite{Sunberg2018} help this, but   rely on  sampling-based methods which  struggle to handle rare events safely. Online solvers also struggle to produce smooth trajectories. 

To address the scalability limitations of POMDP solvers, \cite{Chen2023} construct an ego trajectory tree alongside a scenario tree of multimodal environment predictions, allowing for policy selection that accounts for ego influence on other agents. The approach is extended in~\cite{huang2024} by introducing a differentiable planning framework that jointly optimizes prediction and cost evaluation models to improve planning efficiency.

A similar alternative is contingency planning, where a forward reachable set is used to capture all plausible agent behaviors and to ensure a safe response exists. In~\cite{mustafa2024}, a Bayesian contingency planner is proposed that updates beliefs over agent intentions to refine future responses. Risk-awareness is further integrated into multipolicy contingency planning in~\cite{li2023} to balance safety and efficiency. However, neither these mentioned scenario tree-based methods nor the contingency planners explicitly reason about occlusions.

In this work, we combine occlusion-aware prediction, filter-aware planning, and contingency planning into a unified tree-based framework. This allows us to proactively evaluate possible outcomes, including worst-case interactions with unseen agents, and to generate motion plans that remain safe without relying on overly conservative assumptions. In the following section, we formalize this problem and present our approach.

\section{Problem Formulation}
Trajectory planning for AVs involves computing safe, efficient, and comfortable trajectories that reach a goal. In practice, this is solved iteratively via a receding-horizon: at each step, a trajectory is planned based on the latest observations and prior knowledge, then used as a reference for lower-level vehicle control. Performance evaluation happens retrospectively via a cumulative cost, which models safety, efficiency, and comfort.
Planning requires addressing several interconnected challenges, including perception, state estimation, prediction, trajectory planning, and safety evaluation.  Our main concern in this work lies in planning and reasoning about traffic participants hidden from view, so, we focus on (occupancy) prediction,  planning, and safety w.r.t collision avoidance in the presence of occlusions. To isolate these aspects, we assume perfect perception and state estimation within the ego vehicle's line of sight—  all visible objects are correctly detected and tracked.

Prediction— or forecasting future states of all agents— is particularly challenging when the agents' behavior depends on interactions between all participants. In some cases, even the agents themselves may be uncertain about their next action. Nonetheless, modern predictors  effective at identifying the \textit{most likely} evolution of a scene~\cite{Shi2024}. Thus, many models consider only a single, high-confidence prediction. However, such models struggle with rare events and often do not account for agents emerging from occluded regions~\cite{Yang2023}. To address this, we combine a likely prediction with set-based reachability: the former captures the expected outcome, while the latter, inspired by~\cite{althoff2014, Pek2021}, provides conservative occupancy estimates to ensure safety. For occluded agents, we track and propagate  hidden states, similarly to~\cite{ftu, ptu, Nager2019}.

A second key challenge is planning trajectories that reach desirable states while avoiding unsafe ones. These safety and liveness constraints are critical, but they can significantly restrict optimization. A common approach is to plan a primary trajectory based on a cost function, backed by a static safe fallback. A safety filter or model checker then verifies the primary plan and intervenes if needed. However, as we will show (and as noted in~\cite{hsu2024}), the primary planner must be filter-aware for this approach to be effective. Rather than aiming for general-purpose cost functions and constraints, we define a basic set tailored to the occlusion scenarios studied here.

We solve the trajectory planning problem in real time using a sampling-based, tree-structured planner inspired by~\cite{McNaughton2011, Heinrich2017}. It samples both the most likely outcomes (what we hope for) and a conservative set (what we prepare for), enabling fallback planning under uncertainty. In this work, we consider velocity planning, which is sufficient for the occlusion scenarios studied here, though the approach could be extended to lateral planning as well. This class of planners offers strong versatility and handles complex constraints, though it often incurs computational overhead and yields suboptimal plans. However, we argue that suboptimality is acceptable, as the multi-objective trade-off between safety, comfort, and efficiency is inherently subjective and difficult to define precisely.

\subsection{Problem Definition}
We now formally define the trajectory planning problem. This involves specifying the state space, action space, transition function, cost function, and optimization objective.

\subsubsection{Multi-Agent Transition System}
The trajectory planning problem is defined within the context of a continuous, simultaneous, multi-agent  transition system \( \mathbb{T} \), characterized by the set of agents (I), the set of states that described the environment (S), the set of actions choices for the agents (A), and the transition function that governs the evolution of the environment (T). Formally, 
\[
\mathbb{T} = (I, S, A, T),
\]

\subsubsection{Agents}
The set of agents I consists of the ego vehicle ($\ego$), and the full environment ($\env$), which represents all other visible and hidden vehicles. Formally, \textbf{$I=\{\ego, \env\}$}. 

\subsubsection{State Space}  
The state space, representing all possible observed world states, is defined as:
\[
S \subseteq S^{\ego} \times S^{\env}
\]
where \( S^{\ego} \) is the ego vehicle’s state space, and \( S^{\env} \) represents the state space of the environment observations.
 
\paragraph{Ego Vehicle}
We allow for a flexible state space 
\[
S^{\ego} \subseteq \mathbb{R}^n \times \mathcal{P}(C)
\]
where the state \( s^\ego_t \in S^\ego\) can include $n$ continuous variables in $\mathbb{R}^n$ (e.g., position, velocity, acceleration, jerk) and all relevant categorical attributes via the power set $\mathcal{P}(C)$, where $C$ is a finite label set. For instance, $C$ might include lane indices, vehicle types, or any other discrete labels.

\paragraph{Environment}
The state space of the observations of the environment is given by
\[
S^{\env} \subseteq \mathcal{P}(\mathbb{R}^m \times \mathcal{P}(C))
\]
where \( s^\seen_t \cup s^\hid_t = s^\env_t \in S^\env\) denotes the environment observation state at time $t$,
where:
\begin{itemize}
    \item \( s^\seen_t = \{ y_t^k \}_{k \in K} \) represents the states of \( K \) observed traffic participants, each having $m$ continuous state variables (e.g., position, velocity) and relevant categorical attributes (e.g., vehicle type and lane associations).
    \item \( s^\hid_t  \) represents a set of possible states for participants hidden from the ego vehicle due to occlusions.
\end{itemize}

This decomposition facilitates the modeling of interactions between the ego vehicle, visible participants, and possible hidden participants, which is crucial for ensuring safe and efficient trajectory planning.

\subsubsection{Action Space}\label{actions}
Actions $a_{t:t+\Delta t} \in A$ is factored by the actions of the ego agent ($a_{t:t+\Delta t}^\ego\in A^\ego$) and the actions of the environment ($a_{t:t+\Delta t}^\env\in A^\env$).

\paragraph{Ego Vehicle}
The action space \( A^\ego(s_t^\ego) \) defines all feasible action profiles \( a^\ego_{t:t+\Delta t} \) that the ego vehicle can apply at a given state \( s_t^\ego \) over a stage duration \( \Delta t \).
The state evolves continuously according to
\[
s^\ego_{t'} = f(s_t^\ego, a^\ego_{t:t'}), \forall t' \in [t, t+\Delta t],
\]
and the action profile is said to be feasible subject to the state- and action-dependent constraints if:
\[
\phi(s^\ego_{t'}, a^\ego_{t:t'}) \leq 0, \forall t' \in [t, t+\Delta t].
\]
The action space can be expressed as

\begin{equation}
\begin{aligned}
    A^\ego(s_t^\ego) = \Big\{ a^\ego_{t:t+\Delta t} \mid \; \phi(f(s_t^\ego, a^\ego_{t:t'}), a^\ego_{t:t+\Delta t}) \leq 0, \\
    \forall t' \in [t, t+\Delta &t] \Big\}
\end{aligned}
\end{equation}

\paragraph{Environment}
The action space of the environment \(A^\env(s^\env_t)\) defines all feasible action profiles that every observed or hidden traffic participant may take over the time interval \([t,\,t+\Delta t]\). 
We define it as the (infinite) product
\[
A^\env(s^\env_t) = \prod_{\,y_t \,\in\, s^\env_t}\mathcal{A}(y_t)
\]
where the action space for a participant in a state $y_t \,\in\, s^\env_t$ is
\begin{equation}
\begin{aligned}
\mathcal{A}(y_t)
\;=\;
\Bigl\{
\,a_{t:t+\Delta t}^y \,\Big|\,
\varphi\bigl(g(y_t,\,a_{t:t'}^y),\,a_{t:t'}^y\bigr)\le0,\;
\\ \forall t' \in [t,\, t+\Delta &t]
\Bigr\}
\end{aligned}
\end{equation}
and the participant’s state evolves continuously according to
\begin{equation}
y_{t'} 
\;=\; 
g\bigl(y_t,\;a_{t:t'}^y\bigr),
\quad
\forall\,t' \in [t,\, t+\Delta t],
\end{equation}
subject to feasibility constraints
\begin{equation}
\varphi\!\bigl(y_{t'},\,a_{t:t'}^y\bigr) \;\le\; 0,
\quad
\forall t' \in [t,\, t+\Delta t].
\end{equation}

\subsubsection{Transition Function}
We denote the ego vehicle’s \emph{field of view} by 
\(\FoV\bigl(s^\ego_{t+\Delta t}\bigr)\) and participants \emph{occupancy in space} with $\Occ\bigl(y_{t+\Delta t}\bigr)$. We partition the participants occupancy at time \(t+\Delta t\) into those that end up \emph{inside} this $\FoV$ (i.e. observed) and those that end up \emph{outside} (i.e. hidden).
Accordingly, we define the transition 
\(\,T\colon S \times A \to S\)
by
\begin{equation}
s_{t+\Delta t}
  \;=\;
  T\bigl(s_t,\;a_{t:t+\Delta t}\;\bigr)
  \;=\;
  \Bigl(
    s^\ego_{t+\Delta t}, \;
    s^\seen_{t+\Delta t},\;
    s^\hid_{t+\Delta t}
  \Bigr),
\end{equation}
where $ s^\ego_{t+\Delta t} = f(s_t^\ego, a^\ego_{t+\Delta t}) $ and
\[
  s^\seen_{t+\Delta t}
  \;=\;
  \Bigl\{
    y_{t+\Delta t}
    \;\Big|\;
    \,y_t \in s^\env_t \, ,  \,  y_{t+\Delta t} = g\bigl(y_t,\,a_{t:t+\Delta t}^\env\bigr) \, \land 
  \]
\begin{equation}
    \quad
    \,
    \Occ\bigl(y_{t+\Delta t}\bigr)
    \,\in\,
    \FoV\bigl(s^\ego_{t+\Delta t}\bigr)
  \Bigr\},
\end{equation}
\[
  s^\hid_{t+\Delta t}
  \;=\;
  \Bigl\{
     y_{t+\Delta t}
    \;\Big|\;    
    y_{t+\Delta t} \in \Reach\big(\, g, \, s^\env_t, \, A^\env(s^\env_t), \, \Delta t\big) \, \land 
\]
\begin{equation}
    \Occ\bigl(y_{t+\Delta t}\bigr)
    \,\notin\,
    \FoV\bigl(s^\ego_{t+\Delta t}\bigr)
  \Bigr\}, 
\end{equation}
where the reachable set $r_{t+\Delta t} = \Reach\big(\, g, \, s^\env_t, \, A^\env, \, \Delta t\big)$ is

\[ 
\Reach\big(\, g, \, s^\env_t, A^\env, \Delta t  \big) = 
\Bigl\{
    y_{t+\Delta t}
    \;\Big|\;
    y_t \in s^\env \, ,  \, a^\env_t \in A^\env \, \land  \,
    \]
\begin{equation}
    y_{t+\Delta t} = g\big(y_t,\,a^\env_t \big)
\Bigr\}.
\end{equation}

In other words, every other participant---whether previously seen or hidden---applies one action from its feasible set and transitions to a new state under the dynamics \(g\).  Any participant whose new state lies in the ego's field of view becomes (or remains) part of the new observed set \(s^\seen_{t+\Delta t}\).  Meanwhile, those ending outside the field of view join (or remain in) the new hidden set \(s^\hid_{t+\Delta t}\).  
This construction captures all cases, including previously hidden participant \emph{emerging} into view and previously seen participant becoming occluded.

\subsection{Sampling}
While \( \mathbb{T} \) provides a comprehensive representation of the problem, its infinite state and action spaces make direct optimization computationally intractable. To address this, we construct sampled realizations \( \mathbb{T}' \), defined as:
\begin{equation}
\mathbb{T}' = (S, A', T),
\end{equation}
where $A'\subset A$ are defined through a finite set of sampled actions \( a^\ego_{t:t+\Delta t} \) from \( A^\ego(s_t^\ego) \) and the resulting finite set of states. These realizations approximate the behavior of the original continuous transition system \( \mathbb{T} \) and serve as the basis for solving the trajectory planning problem.

By using \( \mathbb{T}' \), we enable the formulation of policies that govern action selection for the ego vehicle. This concept is formally introduced next.

\subsubsection{Policy Definition}
We define the joint policy as a mapping $(\pi, \mu):S \to (A^\ego, A^\env) = A$.
\paragraph{Ego Vehicle}
A policy \( \pi \) defines how the ego vehicle selects actions based on the observed world state. Formally, it is a mapping, from the joint state space of the ego vehicle and the environment ($S=S^\ego\times S^\env$) to the feasible actions $(a^\ego_{t:t+\Delta t})$ for the ego vehicle as defined in Section~\ref{actions}, i.e., 
\begin{equation}
\pi_{t:t+\Delta t}: S(t) \to A^\ego(s^\ego_t).
\end{equation}

The policy \( \pi \) specifies an action \( a^\ego \in A^\ego \) for any given state \( s \in S \), thereby governing the evolution of the ego vehicle within the continuous transition system \( \mathbb{T} \). The set $\Pi(\mathbb{T})$ notates all possible ego polices $\pi$.
In practice, the policy is applied within a sampled realization \( \mathbb{T}' \), ensuring computational tractability while optimizing performance.

\paragraph{Environment}
The function \( \mu \) is a policy that defines how the environment evolves, which we refer to as an evolution $\mu$. Formally, it is a mapping, 
\begin{equation}
\mu_{t:t+\Delta t}: S(t) \to A^\env(s^\env_t),
\end{equation}
i.e., from the joint state space of the ego vehicle and the environment ($S=S^\ego\times S^\env$) to the feasible actions of the environment, represented by the visible and hidden vehicles defined in Section~\ref{actions}.

The set $M(\mathbb{T})$ defines all possible environmental evolutions $\mu$. Note that although $M(\mathbb{T'}) \subseteq M(\mathbb{T})$, we do not miss any possible action in $A^\env(s_t^\env)$. The subset arises due to our possible observations dependence on the ego actions that we sample and evaluate.

We will single out one specific evolution for the environment, called $\mu^\hope$. In $\mu^\hope$, the action \( a^y_{t:t+\Delta t} \) is chosen such that visible participants follow a predicted trajectory, and any hidden participants remain outside of the ego vehicle's view.

\paragraph{Traces}
Given a transition system $\mathbb{T}$, a trace $w$ is defined by a continuous set of states 
$w_{t_0:t_n} = \{ s_{t'} \mid t' \in [t_0, t_n] \} $. 
The trace $w_{t_0:t_n}$ can  be further decomposed into: 
$w_{t_0:t_n} = (w^\ego_{t_0:t_n}, w^\env_{t_0:t_n})$
where 
$w^\ego_{t_0:t_n} \cap w^\env_{t_0:t_n} = \emptyset$ 
if 
$\Occ(s^\ego_{t'})\cap \Occ(s^\env_{t'})=\emptyset \quad \forall t' \in [t_0,t_n].$ 

We say that an ego policy $\pi$ and an environmental evolution $\mu$ induce a trace $w^{\pi, \mu}_{t_0:t_n} = (w^{\ego, \pi, \mu}_{t_0:t_n}, w^{\env, \pi, \mu}_{t_0:t_n}),$ 
such that:
$s_{t_0}$ 
is the initial state, and
$s_{t+\Delta t}=(T(s_t, \pi(s_t), \mu(s_t)))$ 
for all $t\in[t_0,t_n - \Delta t],$ and any $\, \Delta t < t_n $.

\subsubsection{Cost Function}
The performance of a policy \( \pi \)  under an environmental evolution $\mu$ is evaluated through a cumulative cost function \( J(w^{\pi, \mu}) \), which integrates the instantaneous cost \( c(s, a^\ego) \) over the planning horizon and incorporates a terminal cost \( c_f(s) \) at the end of the horizon:
\begin{equation}
J(w^{\pi, \mu}_{t_0:t_n}) = \int_{t_0}^{t_n} c\big( s_\tau, \pi(s_\tau) \big) \, d\tau + c_f\big(s_{t_n}\big),
\end{equation}
where:
\begin{itemize}
    \item \( s_\tau \): The state at time \( \tau \), evolving under the transition function T with respect to policy $\pi$ and evolution $\mu$.
    \item \( \pi(s_\tau) \): The action selected by the policy \( \pi \) at state \( s_\tau \).
    \item \( c(s, a^\ego) \): The instantaneous cost, quantifying safety, efficiency, and comfort.
    \item \( c_f(s) \): The terminal cost, which evaluates the final state \( s_{t_n} \) at the end of the planning horizon. 
\end{itemize}

This cost function captures both the immediate and long-term effects of the ego vehicle's trajectory. By minimizing \( J(\pi, \mu) \), the trajectory planning problem balances competing objectives, ensuring safety, efficiency, and comfort. 

With the cost function defined, we now formalize the optimization problem as a joint minimization over the policy \( \pi \) and the sampled transition system \( \mathbb{T}' \), ensuring absolute safety over all possible environmental conditions $\mu\in M$.

\subsubsection{Optimization Objective}
The trajectory planning problem is formulated as a joint optimization over sampled transition systems \( \mathbb{T}' \) and policies \( \pi \):
\begin{equation}
\min_{\mathbb{T}' \in \mathbb{T}} \min_{\pi \in \Pi(\mathbb{T}')} J\big(w^{\pi, \mu^\hope}_{t_0:t_n}\big),
\end{equation}
\vspace{-0.75em}
\[ \text{s.t } \quad w^{\ego, \pi, \mu}_{t_0:t_n} \cap w^{\env, \pi, \mu}_{t_0:t_n} = \emptyset, \quad \forall \mu \in M(\mathbb{T}'). \]

This joint optimization synthesizes  a transition system \( \mathbb{T}' \) and a policy \( \pi \) that \textbf{hopes for the best, and prepares for the worst}. The optimization choose a policy that performs the best under the predicted scenario of no new participants emerging from an occlusion, by minimizing  the cost function \( J(\pi) \) over one scenario that is under the predicted evolution $\mu^\hope$. But at the same time, it guarantees that the policy is prepared for all cases, including the worst cases, by enforcing that the policy $\pi$ avoids collisions in \emph{any} evolution $\mu \in M(\mathbb{T}')$. 

In the next section, we show how we sample and use reachability analysis to find a computationally tractable algorithm to solve this problem. By sampling, we allow exploration of alternative realizations of the transition system, and by using reachable sets, we can guarantee absolute collision avoidance.

\section{Method}

Before describing the computational method, we briefly connect it to the problem formulation. The trajectory planning problem requires jointly optimizing over transition systems and policies while ensuring safety in all possible evolutions. Given the complexity of solving this optimization directly, we adopt a sampling-based approach combined with reachability analysis to ensure feasibility. Our method (A) iteratively builds a tree of trajectories, (B) evaluates their validity using both predictive ``hope''-based predictions and conservative ``prepare''-based safety constraints, (C) prunes suboptimal candidates, and (D) selects the lowest-cost feasible trajectory for execution. This approach ensures that we always plan optimistically while preparing for worst-case scenarios.

\subsubsection{Tree Expansion} At a fixed planning frequency, we build a tree of trajectories (or traces) in a breadth-first fashion, level by level from the current state (the root). At each level, we sample new trajectories from each parent state’s endpoint. The first level of trajectories ideally have the length of a planning cycle (or the AVs reaction time), while the other levels can have arbitrary length to yield a suitable planning horizon.

Fig.~\ref{fig:schematic_tree} provides a schematic example of this expansion process, where trajectories are explored, expanded, and evaluated. Trajectories are only retained for further expansion if they meet the hope and prepare conditions, explained next.

\subsubsection{Hope and Prepare Conditions} A trajectory $w^\ego_{t_n:t_{n+1}}$ is considered valid for expansion if it satisfies two conditions: (1) it is \emph{collision-free} with respect to the predicted environment evolution $\mu^\hope$, and (2) there exists a \emph{fallback policy} $\pi^\safe$ from the trajectory end state that ensures safety against all possible future environment evolutions $\mu$. In Fig.~\ref{fig:schematic_tree}, the red nodes indicate that (1) or (2) was not met.

The first condition ensures that we only expand trajectories that are feasible in the expected environment, i.e. if
\begin{equation}
w^\ego_{t_n:t_{n+1}} \cap w^\seen_{t_n:t_{n+1}} \neq \emptyset,
\end{equation}
where $w^\seen_{t_n:t_{n+1}}$ is the trace of the predicted environment evolution $\mu^\hope$.
To perform well in this ``hope'' step, the algorithm relies on three key components: an \emph{environment predictor} $\mu^\hope$ to estimate the most likely evolution of the surroundings, an \emph{action sampler} to generate candidate trajectories $w^\ego_{t_n:t_{n+1}}$, and a \emph{collision checker} to filter out infeasible options.

The second condition guarantees that a \emph{fallback policy} $\pi^\safe$ exists in case any other action is taken by the environment. We define the fallback by the ego trace reaching a safe state at time $t_s$ ($w^{\ego, \pi_\safe}_{t_{n+1}:t_s}$),  the  trace containing all the environment's \emph{reachable states} ($w^{\env}_{t_{n+1}:t_s}$), then  ensure no possible collisions:
\begin{equation}
w^{\ego, \pi_\safe}_{t_{n+1}:t_s} \cap w^{\env}_{t_{n+1}:t_s} \neq \emptyset.
\end{equation}

This step depends on a reachability analysis which over-approximates possible future states to verify the existence of at least one safe continuation. A tighter over-approximation improves the ability to find at least one fallback policy, allowing more ``hope'' trajectories to be retained. 

\subsubsection{Pruning} After generating each level of the tree, we prune to prevent exponential growth in the number of trajectories. Specifically, we partition the state space into $B$ bins and retain only the trajectory with the lowest cumulative cost in each bin. By keeping only the best trajectory in each bin, we maintain  a broad set of feasible options while pruning the tree. In Fig.~\ref{fig:schematic_tree}, the blue circle indicates a node that has been pruned and thus not expanded.

To further reduce computational load, we can defer costly reachable set analysis and collision checks until this pruning phase (e.g., the ``hope'' and ``prepare'' conditions). This avoids unnecessary checks on lower-ranked trajectories.

\subsubsection{Trajectory Selection}
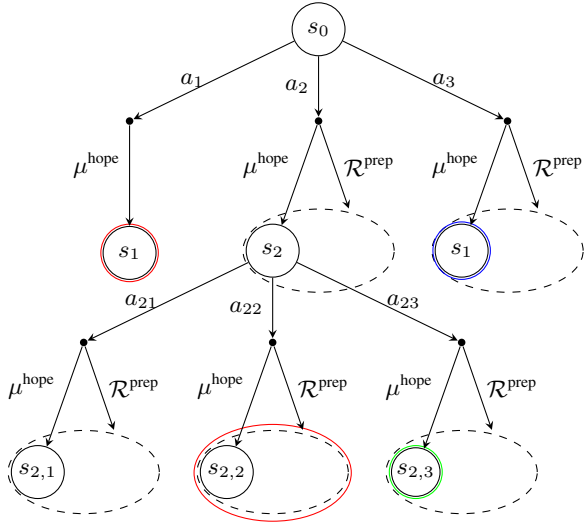
\begin{figure}[t]
\centering
\input{tree}
\caption{An illustration of a minimal sampled transition system. The AV samples candidate actions $a$ and the environment either transitions into a predicted state with $\mu^\hope$ or into any state within the over-approximated reachable set $\mathcal{R}^\text{prep}$. The joint resulting state $s$ is deemed safe (black nodes) if there exists a safe action $\pi(s)$ such that the AVs trace does not intersect \text{any} trace the environment could follow. Unsafe states (red) and pruned states (blue) are not expanded. When the final level is reached, the planned trajectory is found from the leaf node with the lowest cumulative cost (green).}
\label{fig:schematic_tree}
\end{figure}

After pruning, the remaining trajectories form the final expanded tree. When the tree is fully expanded, we select the \emph{leaf node with the lowest cumulative cost} as the best trajectory (illustrated in green in Fig.~\ref{fig:schematic_tree}).
This selection ensures that we execute the most efficient feasible trajectory while maintaining safety constraints. If, at any point, no feasible trajectories remain, a \emph{safe policy is executed immediately from the current state}. Given that the environment evolves as modeled, such a trajectory is always guaranteed to exist, as verified in the previous planning cycle that led to the current state.

Algorithm~\ref{alg:planner} presents the formal pseudocode of our planner, summarizing the steps described above. 
The three nested for loops builds the tree $\mathcal{T}$ by iterating over the $N$ levels, the $B$ best traces, and the $Q$ sampled actions extending each trace. The loop skips and continues \emph{if} the ``hope'' or ``prepare'' conditions are not met. After each completed level, the described pruning function is called, and at the end the best trajectory is returned by finding the best leaf.

The pseudocode follows the notation from our problem formulation while also introducing the tilde notation, which extends operators like $f$, $g$, $T$, and $\Reach$ to generate traces rather than single states. Specifically, while $T(s_t, a_{t:t+\Delta t})$ outputs only the next state $s_{t+\Delta t}$, $\tilde{T}(s_t, a_{t:t+\Delta t})$ recursively applies T, storing intermediate states to construct the full trace $w_{t_0:t_n}$.

\begin{algorithm}
\caption{Hope for the best, prepare for the worst}
\label{alg:planner}
\begin{algorithmic}[1]
\Require Current state, $ s_{t_0} = (\, s^\ego_{t_0}, s^\seen_{t_0}, s^\hid_{t_0} \,)$, goal state, $ s^\text{des}_{t_f} $
\Ensure Planned trajectory, $w^\ego_{t_0:t_N}$

\State $\mathcal{T}_0 \gets \{\, w_{t_0:t_0} = s_{t_0}\, \}$
\vspace{0.1em}
\State $w^{\env}_{t_0:t_s} \gets \tilde{\Reach}\big( g, s^\env_{t_0}, A^\env , t_s\big)$
\For{$n \in [0,\, \ldots, \,N-1]$}
    \State $\mathcal{T}_{n+1} \gets \emptyset$
    \For{$w_{t_0:t_n} \in \mathcal{T}_n$}
    \State $s_{t_n} = (\, s^\ego_{t_n},\, s^\env_{t_n}\,) \gets w_{t_0:t_n}(t_n)$
        \For{$ q \in [0,\, \ldots, \,Q-1]$}
            \State $a^\ego_{t_n:t_{n+1}} \gets \text{sampleAction}(s^\ego_{t_n}, s^\text{des}_{t_f}, q)$
            \State $w^\hope_{t_n:t_{n+1}} \gets \tilde{T}\Big(s_{t_n}, \big(a^\ego_{t_n:t_{n+1}}, \, \mu^\hope_{t_n:t_{n+1}}(s^\env_{t_n})\big)\Big)$
            \State $(\, s^\ego_{t_{n+1}},\, s^\env_{t_{n+1}}\,) \gets w^\hope_{t_n:t_{n+1}}(t_{n+1})$
            \vspace{0.1em}
            \State $w^{\ego}_{t_{n+1}:t_s} \gets \tilde{f}(s^\ego_{t_{n+1}}, \pi^\safe_{t_{n+1}:t_s})$            
            \State $(\, w^\ego_{t_n:t_{n+1}},\, w^\seen_{t_n:t_{n+1}},\, w^\hid_{t_n:t_{n+1}}\,) \gets w^\hope_{t_n:t_{n+1}}$
            \vspace{0.1em}
            \State $w^{\env}_{t_{n+1}:t_s} \gets \tilde{\Reach}\big( g, s^\env_{t_{n+1}}, A^\env , t_s - t_{n+1}\big)$
            \If{$w^\ego_{t_n:t_{n+1}} \cap w^\seen_{t_n:t_{n+1}} \neq \emptyset$}
                \State \textbf{continue} \Comment{No ``hope''}
            \EndIf
            \If{$ w^{\ego, \pi_\safe}_{t_{n+1}:t_s} \cap w^{\env}_{t_{n+1}:t_s} \neq \emptyset$}
                \State \textbf{continue} \Comment{Not ``prepared''}
            \EndIf
            \If{$n=0$}
                \If{$(w^\ego_{t_0:t_1} \cup w^{\ego}_{t_1:t_s}) \cap w^\env_{t_0:t_s} \neq \emptyset$}
                \State \textbf{continue} \Comment{Not safe}
                \EndIf
            \EndIf            
            \State $\mathcal{T}_{n+1} \gets \mathcal{T}_{n+1} \,\cup\, \{\,w_{t_0:t_{n+1}}\}$
        \EndFor
    \EndFor
    \State $\mathcal{T}_{n+1} \gets pruneLevel(\mathcal{T}_{n+1}, B)$
\EndFor

\If{$\mathcal{T}_N \neq \emptyset$}
    \State $w^\ego_{t_0:t_N} \gets getBestTrajectory(\mathcal{T}_N)$
\Else
    \State $w^\ego_{t_0:t_N} \gets \tilde{f}(s^\ego_0, \pi^\safe_{t_0:t_N})$ \Comment{Safety filter}
\EndIf
\State \textbf{return} $w^\ego_{t_0:t_N} \in w_{t_0:t_N}$

\end{algorithmic}
\end{algorithm}

We provide some intuition to the safety guarantees of the system under two assumptions: (1) the ego vehicle starts at $t=0$ in a state that is safe and remain safe, called $\pi_{\safe,0}$ and (2), the reachability calculations are exact, or an over-approximation.  
In all  timesteps $t\geq 1$, an agent samples a series of alternatives with the above tree pruning methodology. If there exists a ``hopeful'' plan that also has a safe fallback trajectory at step $t+1$ under all possible agents’ reachable states ($\pi_{\safe,t+1}$), the agent acts. If not, it proceeds to the safe fallback trajectory determined in timestep $t-1$ ($\pi_{\safe,t}$).  
Thus, at all timesteps under reachability assumptions, a safe trajectory always exists. 
In practice, we compute reachable sets constrained by traffic rules, following the principle in \cite{althoff2014}.
For details on this iterative fail-safe planning approach, see~\cite{Pek2021}.

\section{Experiments}

\begin{figure}[hbt]
\centering
\includegraphics[width=0.99\linewidth,trim={0 16cm 19cm 8cm},]{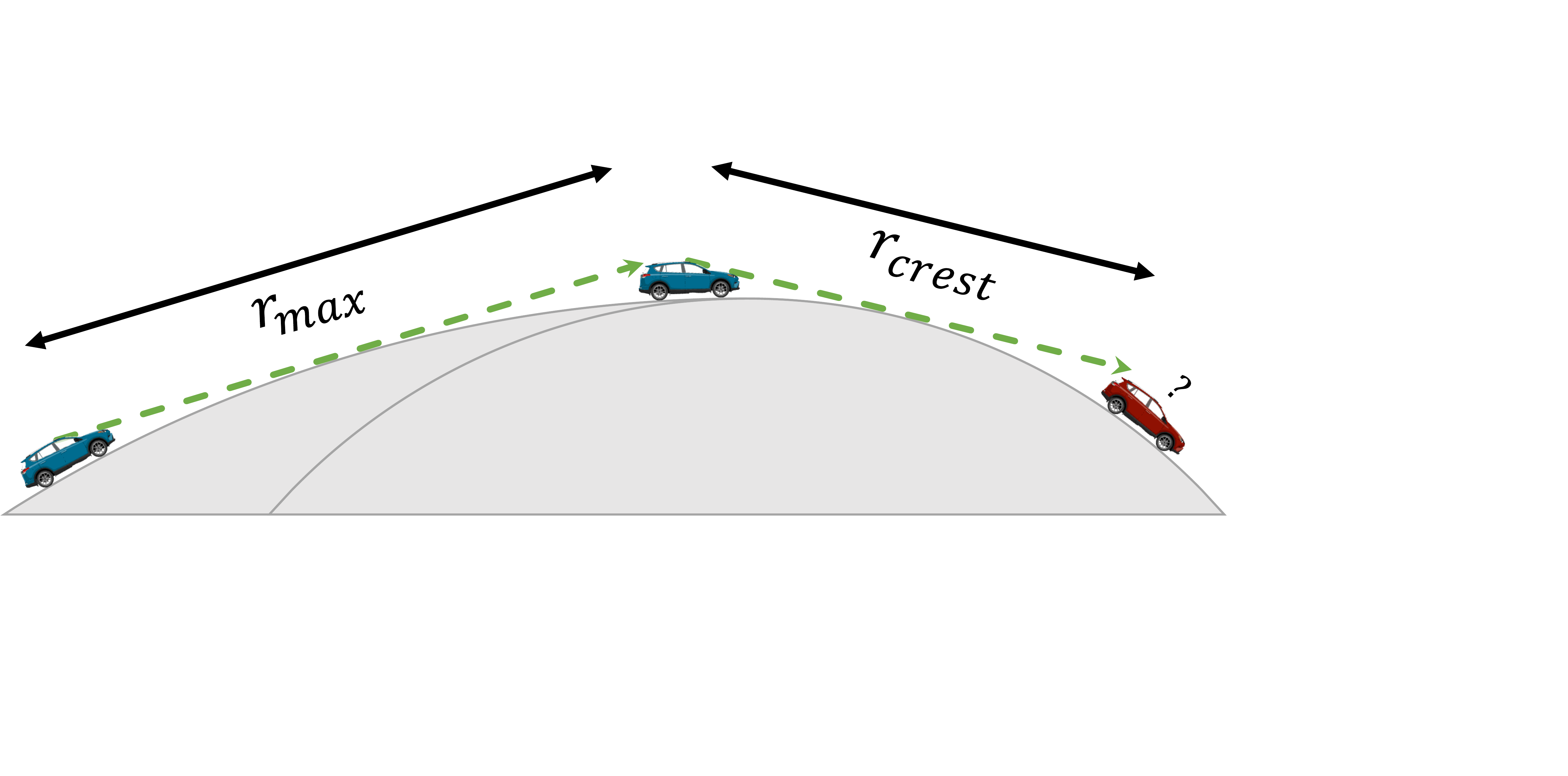}
\caption{The blue AV is traveling along an arched crest. Initially, the sensor range is $r_{max}$, but as the curvature of the arch increases, the range decreases to $r_{crest}$. The AV must plan accordingly, and adjust its speed to be able to stop for the potential red car, hiding ahead.}
\label{fig:scenario_sketch}
\end{figure}

We demonstrate our algorithm in a simple 1-dimensional single lane-following
scenario with occlusions due to limited sensing range ahead, for instance caused by a crest (Fig.~\ref{fig:scenario_sketch}). 

We define the ego vehicle's state space as a longitudinal position along a lane, with the three derivatives velocity, acceleration and jerk:
\begin{equation}
s^\ego(t) = (p(t), v(t), a(t), j(t)) \in \mathbb{R}^4.
\end{equation}

Given target velocity $v_\text{des}$, we define the running cost as:
\begin{equation}
c(s(t)) = j(t)^2 + \lambda(v(t)-v_\text{des})^2.
\end{equation}
Although the exact terminal cost can be found by formulating the problem as a linear-quadratic-regulator and solving the algebraic Ricatti equation, we skip this step for simplicity as the planning horizon $t_N$ iss be long enough to model the crest.

We restrict the action space to trajectories with velocities $v(t) \in [0, v^\ego_\text{max}]$ and accelerations $a(t) \in [a^\ego_\text{min}, a^\ego_\text{max}]$ and build our trajectory tree by connecting $N$ quartic polynomials, i.e., $p(t) = \beta_4 t^4 + \beta_3 t^3 + \beta_2 t^2 + \beta_1 t + \beta_0$, as suggested in \cite{werling_2010} for velocity keeping.
The polynomial coefficients, $\beta_1,\beta_2,\beta_3,\beta_4$, are uniquely defined by initial constraints $\big(\,p(t_0), v(t_0), a(t_0)\,\big)$ given by the previous trajectory's end state, and a sampled end constraint $\big( \,v_f, a_f=0 \, \big)$ at a sampled end time $t_f$. This results in optimal longitudinal trajectories where  the coefficients and the trajectory cost can be computed efficiently in closed form~\cite{werling_2010}. We sample $Q$ pairs of $\big( \,v_f, \,t_f \, \big)$ where half of them have linear spaced velocities in $[0, v^\ego_\text{max}]$ and the other half have $v_f = v_{des}$ at linear spaced time horizons. Feasible time horizons are computed from the state and the accelerations bounds. Note that the horizon $t_f$ is independent of the trajectory segment horizon ($t_N/N$, for uniform segment horizons). If the desired state (desired velocity and zero acceleration) can be reached earlier, the trajectory segment is concatenated with a zero acceleration polynomial.

Any other traffic participant, indexed $k$ in the environment, is assumed to be ahead of the ego vehicle and follow similar double integrator dynamics with bounded velocities $v^k(t) \in [0, v^\env_\text{max}]$ and accelerations $a^k(t) \in [a^\env_\text{min}, a^\env_\text{max}]$, where $a^\env_\text{min} < a^\ego_\text{min}$ results in the ego vehicle needing to keep some safe distance (due to having a longer stopping distance).

We assume the ego vehicle can detect any other participant within its current sensor range, $r(s(t))$. 
Positions greater than $r(s^\ego(t)) + p^\ego(t)$ are occluded and outside ego's field of view, $\FoV$. 
To demonstrate our algorithm, we consider a scenario where the sensor range is limited due to the 
(known) road topology ahead. We define the sensor range to be 

\begin{equation}
r(s) = 
\begin{cases} 
\text{min}( \, p(t) + r_\text{max}, \, p_\text{crest} \, ), & \text{if} \quad p(t) < p_\text{crest} - r_\text{crest} \\
p(t) + r_\text{crest}, & \text{else},
\end{cases}
\end{equation}
i.e., the sensor range is initially some $r_\text{max}$, but decreases linearly to $r_\text{crest}$ as the ego vehicle approaches a crest at $p_\text{crest}$. At $ p^\ego \geq p_\text{crest} - r_\text{crest}$, the range is constant at a reduced range, as the ground the vehicle is driving on follows an arc where the road occludes itself. 

In this scenario, we simply pick $\mu^\hope$ to be a policy with constant velocity for any seen vehicle. If the ego vehicle's front position at any point in time is greater than the closest other participant's end position, a collision occurs.

We chose the fallback policy $\pi_\safe$ to after a delay of one planning cycle be braking with the maximum deceleration, bringing the ego vehicle to zero velocity at some earliest possible time $t_{v=0}$. We collision check this against the states at time $t_{v=0}$ in all evolutions $\mu \in M$, which are over-approximated by computing the union of the reachable intervals of positions at the time $t_{v=0}$. For seen participants, the computations are made with their observed velocity $v^k(t_0)$ at $t_0$ and their minimum and maximum accelerations. For the occluded possible participants we compute the reachable sets conservatively for simplicity with both the minimum and maximum velocities and accelerations since no velocity is observed in the occlusions (and we only implement tracked \emph{possible positions} in this work). 

We simulate a scenario with two variants: one where the road ahead is free, and one where a stationary vehicle is hidden behind the crest. The first scenario demonstrates how our planner performs in the best case. The second scenario with the hidden vehicle illustrates the importance of ensuring that the AV can stop if a hidden stationary vehicle appears just outside its field of view. The scenario is illustrated in Fig.~\ref{fig:scenario_sketch} and Tab.~\ref{tab:parameters} summarizes the relevant parameters.

\begin{table}[tb]
    \centering
    \caption{Parameter definitions and values}
    \renewcommand{\arraystretch}{1.3}
    \begin{tabular}{l l l l}
        \hline
        \textbf{Symbol} & \textbf{Description} & \textbf{Value} & \textbf{Units} \\
        \hline
        $v_0$  & Initial velocity & 14.0  & m/s \\
        $v_{des}$ & Target velocity & 15.0  & m/s \\
        $\lambda$ & Velocity cost weight & 1.0   & — \\
        \hline 
        $t_p$ & Re-planning cycle time & 0.2   & s \\
        $t_N$ & Planning horizon & 5.0   & s \\
        $N$ & Trajectory levels & 5     & — \\
        $B$ & Velocity bins & 32    & — \\
        $Q$ & Parameter & 128   & —  \\
        \hline
        $v^\ego_{\max}$ & Ego maximum velocity & 15    & m/s \\
        $[a^\ego_{\min}, a^\ego_{\max}]$ & Ego acceleration limits & $[-5, 3]$  & m/s\textsuperscript{2} \\
        $v^\env_{\max}$ & Environment maximum velocity & 20    & m/s \\
        $[a^\env_{\min}, a^\env_{\max}]$ & Environment acceleration limits & $[-8, 3]$  & m/s\textsuperscript{2} \\ 
        \hline
        $[r_{\text{slope}}, \, r_{\max}]$ & Sensor range & $[15, 50]$ & m \\
        $[p_{\text{slope}}, \,p_{\text{crest}}$] & Crest positions & $[35, 50]$ & m \\
        \hline
    \end{tabular}
    \label{tab:parameters}
\end{table}

\begin{figure}[htb]
\centering
\includegraphics[width=0.99\linewidth]{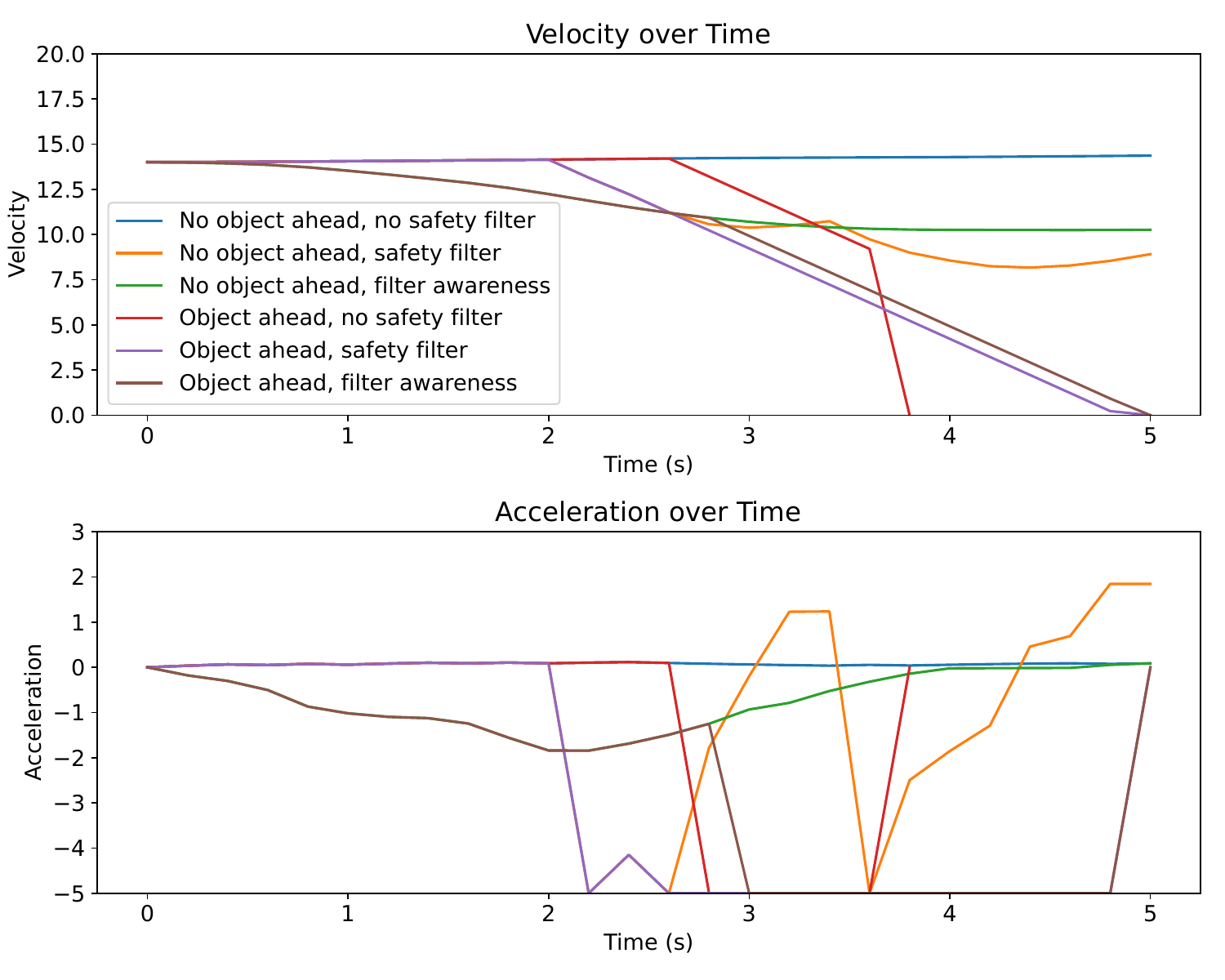}
\caption{Velocity and acceleration for the three planners in the two scenarios.}
\label{fig:results}
\end{figure}

Fig.~\ref{fig:results} shows the resulting state trajectory for an AV planning its motion with three variants of the same planner. The first variant has no occlusion-awareness and only checks its candidate trajectories against seen obstacles and the constant velocity predictions (the ``hope'' condition). This makes it follow the desired velocity closely, however, at the cost of not being able to stop in time when the stationary vehicle gets revealed. The second variant is the same planner, but with the addition of a safety filter which checks collisions against possible hidden objects and their reachable set predictions, similarly to previous work in~\cite{Nager2019, Orzechowski2018, Koschi2021, ftu}. During the first 2s of the scenario, the filter does not alter the AVs planned path, but then it kicks in and initiates an emergency braking maneuver, regardless if an object is ahead or not. If an object is present, the AV can stop just in time, but if it is not, the AV will accelerate rapidly until it again gets restricted by the safety filter. 
The third variant of the planner is our proposed occlusion-aware method, presented in Algorithm~\ref{alg:planner}. Pro-actively, it reduces the AVs speed, removing the need for the safety filter to intervene. However, if an object ahead is detected, the AV is able to stop in time, as illustrated in the scenario where a hidden stationary vehicle is revealed. 

\section{Conclusion and Future Work}
\label{sec:conclusions}
This paper presents an occlusion-aware contingency planner for autonomous vehicles, integrating tree-based motion planning with reachability analysis. Our approach ensures safety by systematically accounting for possible hidden traffic participants while enabling efficient planning by reasoning over both future observations and their absence. The experimental results demonstrate that our approach eliminates sudden interruptions by seamlessly integrating contingency planning within the motion planner, ensuring smooth and predictable behavior while maintaining collision avoidance guarantees.

Future work will focus on extending the method to  settings with multimodal, ego-conditioned predictions. Instead of ``hoping'' for a single predicted evolution, we aim to incorporate multiple plausible evolutions, weighted according to learned probability distributions. 
An expectation operator in the optimization problem could be introduced to account for this, naturally extending the approach toward stochastic dynamic programming.
This directly relaxes the simplified assumption of known behavior by reasoning over multiple plausible futures.

To relax the assumption of perfect perception, future work could explore shrinking the field of view by a small margin so that also visible regions are conservatively treated as potentially occluded. The guarantees would then only hold if these inflated occluded sets are true over-approximations. Additionally, to reduce the computational cost of evaluating multiple reachable sets, one could learn an approximation function that predicts how the safety filter would behave at future timesteps. Applying the exact safety filter at the current state, and the learned approximation at later states, could result in faster forward planning without compromising guarantees.

The trajectory sampling distribution and cost function could also be learned, enabling the planner to better align decisions with human driving preferences. 
With this, the system can optimize for not only safety and efficiency but also comfort and naturalistic driving patterns.
These advancements will further improve the applicability of our method to real-world driving scenarios, bridging the gap between formal safety guarantees and practical decision-making under uncertainty.

\bibliographystyle{IEEEtran}
\balance
\bibliography{references}

\end{document}

%% file: tree.tex
\begin{tikzpicture}[>=stealth, font=\small]

\node[circle, draw, minimum size=0.7cm] (s0) {$s_0$};

\node[fill=black, circle, inner sep=1pt]
      (a1) [below=0.8cm of s0, xshift=-2.5cm] {};
\draw[->] (s0) -- node[midway,left] {$a_1$} (a1);

\node[circle, draw, inner sep=1pt, minimum size=0.7cm]
      (circA1) [below=1.33cm of a1]{$s_{1}$};
\node[circle, draw, inner sep=1pt, minimum size=0.76cm, color=red]
      (circA1r) [below=1.3cm of a1]{};
\draw[->] (a1) -- node[yshift=4pt,left] {$\mu^{\text{hope}}$} (circA1);

\node[fill=black, circle, inner sep=1pt]
      (a2) [below=0.8cm of s0] {};
\draw[->] (s0) -- node[midway,left] {$a_2$} (a2);

\node[ellipse, draw, dashed, minimum width=2cm, minimum height=1.1cm]
      (ovalA2) [below=1.1cm of a2] {};
\node[circle, draw, inner sep=1pt, minimum size=0.7cm]
      (circA2) at ([xshift=0.4cm] ovalA2.west)
      {$s_{2}$};

\draw[->] (a2) -- node[yshift=0pt,right]  {$\mathcal{R}^{\text{prep}}$} ([xshift=0.4cm] ovalA2.north);
\draw[->] (a2) -- node[yshift=4pt,left] {$\mu^{\text{hope}}$} (circA2);

\node[fill=black, circle, inner sep=1pt]
    (a21) [below=0.8cm of circA2, xshift=-2.5cm] {};
\draw[->] (circA2) -- node[midway,left] {$a_{21}$} (a21);

\node[ellipse, draw, dashed, minimum width=2cm, minimum height=1.1cm]
    (oval21) [below=1.1cm of a21] {};
\node[circle, draw, inner sep=1pt, minimum size=0.7cm]
    (circ21) at ([xshift=0.4cm] oval21.west) {$s_{2,1}$};

\draw[->] (a21) -- node[yshift=0pt,right]  {$\mathcal{R}^{\text{prep}}$} ([xshift=0.4cm] oval21.north);
\draw[->] (a21) -- node[yshift=4pt,left] {$\mu^{\text{hope}}$} (circ21);

\node[fill=black, circle, inner sep=1pt]
    (a22) [below=0.8cm of circA2, xshift=0cm] {};
\draw[->] (circA2) -- node[midway,left] {$a_{22}$} (a22);

\node[ellipse, draw, minimum width=2.08cm, minimum height=1.28cm, color=red]
    (oval22r) [below=1.02cm of a22] {};
\node[ellipse, draw, dashed, minimum width=2cm, minimum height=1.1cm]
    (oval22) [below=1.1cm of a22] {};    
\node[circle, draw, inner sep=1pt, minimum size=0.70cm]
    (circ22) at ([xshift=0.4cm] oval22.west) {$s_{2,2}$};        
       
\draw[->] (a22) -- node[yshift=4pt,left] {$\mu^{\text{hope}}$} (circ22);
\draw[->] (a22) -- node[yshift=0pt,right]  {$\mathcal{R}^{\text{prep}}$} ([xshift=0.4cm] oval22.north);

\node[fill=black, circle, inner sep=1pt]
    (a23) [below=0.8cm of circA2, xshift=2.5cm] {};
\draw[->] (circA2) -- node[midway,right] {$a_{23}$} (a23);

\node[ellipse, draw, dashed, minimum width=2cm, minimum height=1.1cm]
    (oval23) [below=1.1cm of a23] {};
\node[circle, draw, inner sep=1pt, minimum size=0.7cm, color=green]
    (circ23) at ([xshift=0.4cm] oval23.west) {};

\node[circle, draw, inner sep=1pt, minimum size=0.6cm]
    (circ23) at ([xshift=0.4cm] oval23.west) {$s_{2,3}$};

\draw[->] (a23) -- node[yshift=0pt,right]  {$\mathcal{R}^{\text{prep}}$} ([xshift=0.4cm] oval23.north);
\draw[->] (a23) -- node[yshift=4pt,left] {$\mu^{\text{hope}}$} (circ23);

\node[fill=black, circle, inner sep=1pt]
      (a3) [below=0.8cm of s0, xshift=2.5cm] {};
\draw[->] (s0) -- node[midway,right] {$a_3$} (a3);

\node[ellipse, draw, dashed, minimum width=2cm, minimum height=1.1cm]
      (ovalA3) [below=1.1cm of a3] {};
\node[circle, draw, inner sep=1pt, minimum size=0.7cm]
      (circA3) at ([xshift=0.4cm] ovalA3.west){$s_{1}$};
\node[circle, draw, inner sep=1pt, minimum size=0.76cm, color=blue]
      (circA3g) at ([xshift=0.4cm] ovalA3.west){};      

\draw[->] (a3) -- node[yshift=0pt,right]  {$\mathcal{R}^{\text{prep}}$} ([xshift=0.4cm] ovalA3.north);
\draw[->] (a3) -- node[yshift=4pt,left] {$\mu^{\text{hope}}$} (circA3);

\end{tikzpicture}

%% file: references.bib
@inproceedings{werling_2010,
  author       = {Werling, Moritz and Ziegler, Julius and Kammel, Sören and Thrun, Sebastian},
  title        = {Optimal trajectory generation for dynamic street scenarios in a Frenét Frame}, 
  booktitle    = {Proc. IEEE Int. Conf. Robot. Autom. (ICRA)},
  pages        = {987-993},  
  year         = {2010},
  doi          = {10.1109/ROBOT.2010.5509799}
}

@misc{dmv2023,
  author       = {{California Department of Motor Vehicles}},
  title        = {Autonomous Vehicle Disengagement Reports},
  howpublished = {California DMV},
  month        = {Dec.},
  year         = {2023},
  note         = {Accessed: Dec. 12, 2024},
  url          = {https://www.dmv.ca.gov/portal/vehicle-industry-services/autonomous-vehicles/disengagement-reports/}
}

@misc{nhtsa2023,
  author       = {{National Highway Traffic Safety Administration}},
  title        = {Incident Report Data Summary},
  howpublished = {NHTSA},
  month        = {Oct.},
  year         = {2024},
  note         = {Accessed: Dec. 12, 2024},
  url          = {https://www.nhtsa.gov/laws-regulations/standing-general-order-crash-reporting}
}

@article{stu,
  author      = {Truls Nyberg and José Manuel Gaspar Sánchez and Vandana Narri and Henrik Pettersson and Jonas Mårtensson and Karl H. Johansson and Martin Törngren and Jana Tumova},
  title       = {Share the Unseen: Sequential Reasoning About Occlusions Using Vehicle-to-Everything Technology},
  journal     = {IEEE Trans. Control Syst. Technol.},
  volume      = {},
  number      = {},
  pages       = {1--14},
  month       = {Nov.},
  year        = {2024},
  doi         = {10.1109/TCST.2024.3499832}
}

@inproceedings{ptu,
  author      = {Truls Nyberg and Jonne van Haastregt and Jana Tumova},
  title       = {Highway-Driving with Safe Velocity Bounds on Occluded Traffic},
  booktitle   = {Proc. IEEE Int. Conf. Robot. Autom. (ICRA)},
  pages       = {6828--6835},
  year        = {2024},
  doi         = {10.1109/ICRA57147.2024.10610904}
}

@inproceedings{ftu,
  author      = {José Manuel Gaspar Sánchez and Truls Nyberg and Christian Pek and Jana Tumova and Martin Törngren},
  title       = {Foresee the Unseen: Sequential Reasoning About Hidden Obstacles for Safe Driving},
  booktitle   = {Proc. IEEE Intell. Veh. Symp. (IV)},
  pages       = {255--264},
  year        = {2022},
  doi         = {10.1109/IV51971.2022.9827171}
}

@article{koopman2023,
  author    = {Philip Koopman},
  title     = {{UL 4600}: What to Include in an Autonomous Vehicle Safety Case},
  journal   = {IEEE Comput.},
  volume    = {56},
  number    = {5},
  pages     = {101--104},
  month     = {May},
  year      = {2023}
}

@article{hsu2024,
  author    = {Kai-Chieh Hsu and Haimin Hu and Jaime F. Fisac},
  title     = {The Safety Filter: A Unified View of Safety-Critical Control in Autonomous Systems},
  journal   = {Annu. Rev. Control Robot. Auton. Syst.},
  volume    = {7},
  pages     = {47--72},
  month     = {July},
  year      = {2024},
  doi       = {10.1146/annurev-control-071723-102940}
}

@article{althoff2014,
  author    = {Matthias Althoff and John M. Dolan},
  title     = {Online Verification of Automated Road Vehicles Using Reachability Analysis},
  journal   = {IEEE Trans. Robot.},
  volume    = {30},
  number    = {4},
  pages     = {903--918},
  year      = {2014},
  doi       = {10.1109/TRO.2014.2312453}
}

@article{RSS,
  author    = {Maria Soledad Elli and Jack Weast},
  title     = {Towards a Formal Model for Safe and Scalable Automated Vehicle Decision-Making: A Brief Survey on Responsibility-Sensitive Safety},
  journal   = {SAE Int. J. Connect. Autom. Veh.},
  volume    = {4},
  number    = {1},
  pages     = {9--22},
  month     = {Mar.},
  year      = {2021},
  doi       = {10.4271/12-04-01-0002}
}

@ARTICLE{Pek2021,
  author={Pek, Christian and Althoff, Matthias},
  journal={IEEE Transactions on Robotics}, 
  title={Fail-Safe Motion Planning for Online Verification of Autonomous Vehicles Using Convex Optimization}, 
  year={2021},
  volume={37},
  number={3},
  pages={798-814},
  doi={10.1109/TRO.2020.3036624}}

@article{Trauth2023,
  author    = {Rainer Trauth and Korbinian Moller and Johannes Betz},
  title     = {Toward Safer Autonomous Vehicles: Occlusion-Aware Trajectory Planning to Minimize Risky Behavior},
  journal   = {IEEE Open J. Intell. Transp. Syst.},
  volume    = {4},
  pages     = {929--942},
  month     = {Nov.},
  year      = {2023},
  doi       = {10.1109/OJITS.2023.3336464}
}

@article{Koschi2021,
  author    = {Markus Koschi and Matthias Althoff},
  title     = {Set-Based Prediction of Traffic Participants Considering Occlusions and Traffic Rules},
  journal   = {IEEE Trans. Intell. Veh.},
  volume    = {6},
  number    = {2},
  pages     = {249--265},
  year      = {2021},
  doi       = {10.1109/TIV.2020.3017385}
}

@inproceedings{Nager2019,
  author    = {Yannik Nager and Andrea Censi and Emilio Frazzoli},
  title     = {What Lies in the Shadows? {Safe} and Computation-Aware Motion Planning for Autonomous Vehicles Using Intent-Aware Dynamic Shadow Regions},
  booktitle = {Proc. IEEE Int. Conf. Robot. Autom. (ICRA)},
  pages     = {5800--5806},
  year      = {2019},
  doi       = {10.1109/ICRA.2019.8793557}
}

@inproceedings{Orzechowski2018,
  author    = {Piotr F. Orzechowski and Annika Meyer and Martin Lauer},
  title     = {Tackling Occlusions and Limited Sensor Range with Set-Based Safety Verification},
  booktitle = {Proc. IEEE Intell. Transp. Syst. Conf. (ITSC)},
  pages     = {1729--1736},
  year      = {2018},
  doi       = {10.1109/ITSC.2018.8569332}
}

@article{Hu2022,
  author    = {Haimin Hu and Kensuke Nakamura and Jaime F. Fisac},
  title     = {{SHARP}: Shielding-Aware Robust Planning for Safe and Efficient Human-Robot Interaction},
  journal   = {IEEE Robot. Autom. Lett.},
  volume    = {7},
  number    = {2},
  pages     = {5591--5598},
  year      = {2022},
  doi       = {10.1109/LRA.2022.3155229}
}

@article{Hu2024,
  author    = {Haimin Hu and David Isele and Sangjae Bae and Jaime F. Fisac},
  title     = {Active Uncertainty Reduction for Safe and Efficient Interaction Planning: A Shielding-Aware Dual Control Approach},
  journal   = {Int. J. Robot. Res.},
  volume    = {43},
  number    = {9},
  pages     = {1382--1408},
  year      = {2024},
  doi       = {10.1177/02783649231215371}
}

@inproceedings{Chen2023,
  author    = {Yuxiao Chen and Peter Karkus and Boris Ivanovic and Xinshuo Weng and Marco Pavone},
  title     = {Tree-Structured Policy Planning with Learned Behavior Models},
  booktitle = {Proc. IEEE Int. Conf. Robot. Autom. (ICRA)},
  pages     = {7902--7908},
  year      = {2023},
  doi       = {10.1109/ICRA48891.2023.10161419}
}

@inproceedings{Huang2024,
  author    = {Zhiyu Huang and Peter Karkus and Boris Ivanovic and Yuxiao Chen and Marco Pavone and Chen Lv},
  title     = {{DTPP}: Differentiable Joint Conditional Prediction and Cost Evaluation for Tree Policy Planning in Autonomous Driving},
  booktitle = {Proc. IEEE Int. Conf. Robot. Autom. (ICRA)},
  pages     = {6806--6812},
  year      = {2024},
  doi       = {10.1109/ICRA57147.2024.10610550}
}

@article{Li2023,
  author    = {Tong Li and Lu Zhang and Sikang Liu and Shaojie Shen},
  title     = {{MARC}: Multipolicy and Risk-Aware Contingency Planning for Autonomous Driving},
  journal   = {IEEE Robot. Autom. Lett.},
  volume    = {8},
  number    = {10},
  pages     = {6587--6594},
  year      = {2023},
  doi       = {10.1109/LRA.2023.3310431}
}

@article{Mustafa2024,
  author    = {Khaled A. Mustafa and Daniel Jarne Ornia and Jens Kober and Javier Alonso-Mora},
  title     = {{RACP}: Risk-Aware Contingency Planning with Multi-Modal Predictions},
  journal   = {IEEE Trans. Intell. Veh.},
  pages     = {1--16},
  year      = {2024},
  doi       = {10.1109/TIV.2024.3411530}
}

@inproceedings{Hubmann2019,
  author    = {Constantin Hubmann and Nils Quetschlich and Jens Schulz and Julian Bernhard and Daniel Althoff and Christoph Stiller},
  title     = {A {POMDP} maneuver planner for occlusions in urban scenarios},
  booktitle = {Proc. IEEE Intell. Veh. Symp. (IV)},
  pages     = {2172--2179},
  year      = {2019},
  doi       = {10.1109/IVS.2019.8814179}
}

@article{Ye2017,
  author    = {Nan Ye and Adhiraj Somani and David Hsu and Wee Sun Lee},
  title     = {{DESPOT}: Online {POMDP} planning with regularization},
  journal   = {J. Artif. Intell. Res.},
  volume    = {58},
  number    = {1},
  pages     = {231--266},
  month     = {Jan.},
  year      = {2017}
}

@inproceedings{Sunberg2018,
  author    = {Zachary Sunberg and Mykel Kochenderfer},
  title     = {Online algorithms for {POMDPs} with continuous state, action, and observation spaces},
  booktitle = {Proc. Int. Conf. Automated Planning and Scheduling (ICAPS)},
  volume    = {28},
  number    = {1},
  pages     = {259--263},
  year      = {2018},
  month     = {Jun.},
  doi       = {10.1609/icaps.v28i1.13882}
}

@phdthesis{McNaughton2011,
  author  = {Matthew McNaughton},
  title   = {Parallel Algorithms for Real-Time Motion Planning},
  school  = {Carnegie Mellon Univ.},
  address = {Pittsburgh, PA},
  year    = {2011}
}

@book{Heinrich2017,
  author    = {Steffen Heinrich},
  title     = {Planning Universal On-Road Driving Strategies for Automated Vehicles},
  publisher = {Springer Vieweg},
  address   = {Wiesbaden, Germany},
  year      = {2017},
  series    = {AutoUni -- Schriftenreihe},
  volume    = {119},
  doi       = {10.1007/978-3-658-21954-3}
}

@inproceedings{Yang2023,
  author    = {Yi Yang and Qingwen Zhang and Thomas Gilles and Nazre Batool and John Folkesson},
  title     = {{RMP}: A random mask pretrain framework for motion prediction},
  booktitle = {Proc. IEEE Int. Conf. Intell. Transp. Syst. (ITSC)},
  pages     = {3717--3723},
  year      = {2023},
  doi       = {10.1109/ITSC57777.2023.10422522}
}

@article{Shi2024,
  author    = {Shaoshuai Shi and Li Jiang and Dengxin Dai and Bernt Schiele},
  title     = {{MTR++}: Multi-agent motion prediction with symmetric scene modeling and guided intention querying},
  journal   = {IEEE Trans. Pattern Anal. Mach. Intell.},
  volume    = {46},
  number    = {5},
  pages     = {3955--3971},
  month     = {May},
  year      = {2024},
  doi       = {10.1109/TPAMI.2024.3352811}
}

@inproceedings{bouton_scalable_2018,
  author    = {Maxime Bouton and Alireza Nakhaei and Kikuo Fujimura and Mykel J. Kochenderfer},
  title     = {Scalable decision making with sensor occlusions for autonomous driving},
  booktitle = {Proc. IEEE Int. Conf. Robot. Autom. (ICRA)},
  pages     = {2076--2081},
  year      = {2018},
  month     = {May},
  doi       = {10.1109/ICRA.2018.8460914}
}
